\documentclass[10pt,twocolumn,letterpaper]{article}

\usepackage{cvpr}
\usepackage{times}
\usepackage{epsfig}
\usepackage{graphicx}
\usepackage{amsmath}
\usepackage{amssymb}
\usepackage{times}
\usepackage{epsfig}
\usepackage{graphicx}
\usepackage{amsmath}
\usepackage{amssymb}
\usepackage{multirow}
\usepackage{bbding}
\usepackage{pifont}
\usepackage[caption=false,font=footnotesize]{subfig}

\usepackage[pagebackref=true,breaklinks=true,letterpaper=true,colorlinks,bookmarks=false]{hyperref}

\cvprfinalcopy 

\def\cvprPaperID{1076} 

\ifcvprfinal\fi
\begin{document}

\title{{FSSD}: Feature Fusion Single Shot Multibox Detector }

\author{Zuo-Xin Li$^1$~~~~~~~~~~Lu Yang$^2$ ~~~~~~~~~~~~~ Fu-Qiang Zhou$^1$\\
	$^1$Key Laboratory of Precision Opto-mechatronics Technology, Ministry of Education,\\
	 Beihang University, Beijing 100191, China\\
   $^2$Beijing University of Posts and Telecommunications\\
	{\tt\small $^1$\{lizuoxin, zfq\}@buaa.edu.cn, $^2$soeaver@bupt.edu.cn}
}

\maketitle


\begin{abstract}
	SSD (Single Shot Multibox Detetor) is one of the best object detection algorithms with both high accuracy and fast speed. However, SSD's feature pyramid  detection method makes it hard to fuse the features from different scales. In this paper, we proposed FSSD (Feature Fusion Single Shot Multibox Detector), an enhanced SSD with a novel and lightweight feature fusion module which can improve the performance significantly over SSD with just a little speed drop. In the feature fusion module, features from different layers with different scales are concatenated together, followed by some down-sampling blocks to generate new  feature pyramid, which will be fed to multibox detectors to predict the final detection results. On the Pascal VOC 2007 test, our network can achieve 82.7 mAP (mean average precision) at the speed of 65.8 FPS (frame per second) with the input size 300$\times$300 using a single Nvidia 1080Ti GPU. In addition, our result on COCO is also better than the conventional SSD with a large margin. Our FSSD outperforms a lot of state-of-the-art object detection algorithms in both aspects of accuracy and speed. Code is available at \url{https://github.com/lzx1413/CAFFE_SSD/tree/fssd}.
\end{abstract}

\section{Introduction}
\begin{figure}[t]
	\centering
	\includegraphics[width=0.8\linewidth]{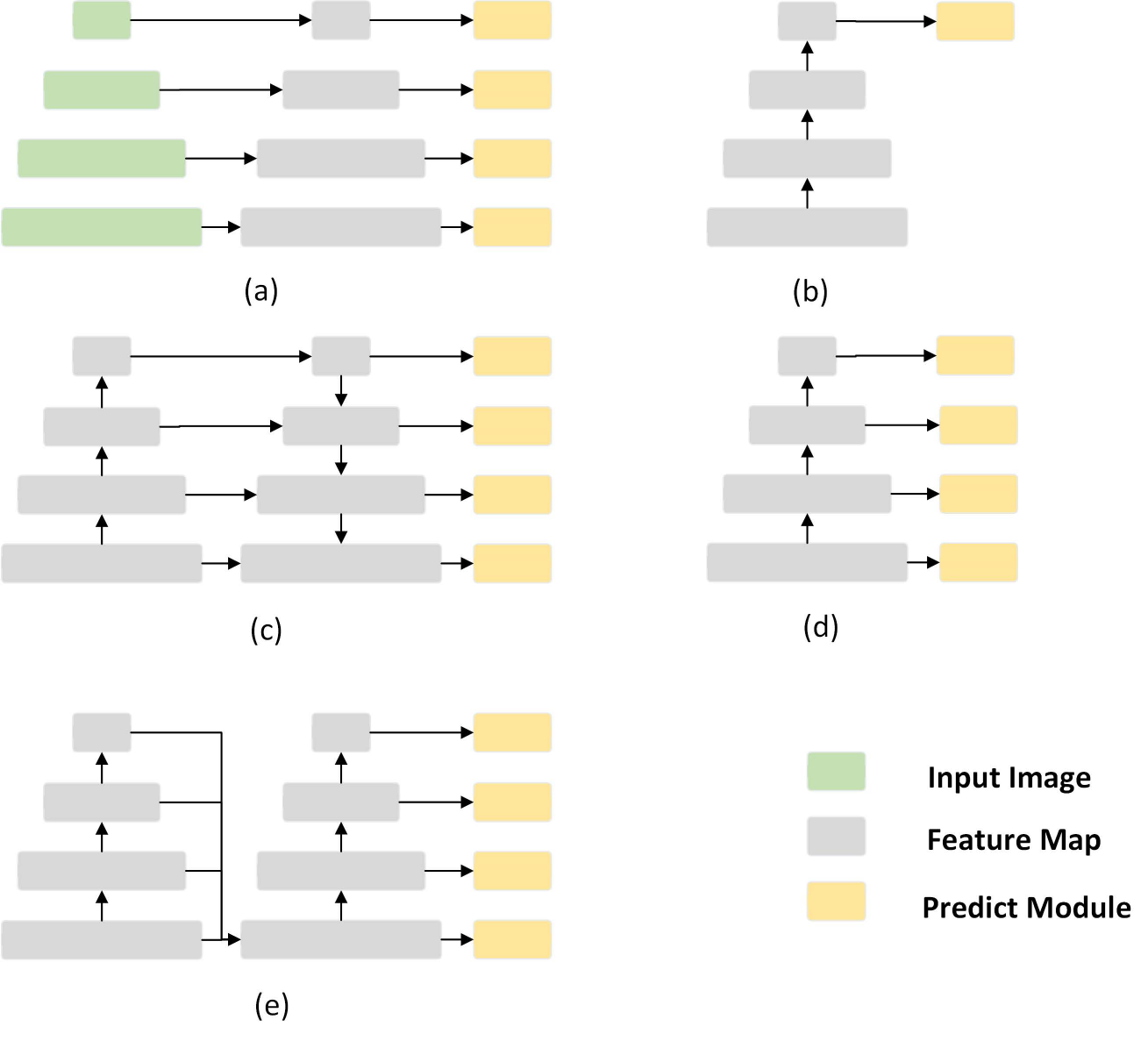}
	\caption{(a) Features are computed from images with different scales independently, which is an inefficient way. (b) Just one scale features are used to detect objects, which is used in some two stage detectors such as Faster R-CNN \cite{FasterRCNN} and R-FCN \cite{RFCN}. (c) Feature fusion method adopted by \cite{FPN,SharpMask}, features are fused from top to bottom layer by layer. (d) Use the feature pyramid generated  from a ConvNet. The conventional SSD is one of the examples. (e) Our proposed feature fusion and feature pyramid generation method. Features from different layers with different scales are concatenated together first and used to generate a series of pyramid features later. }
	\label{fig:fm}
\end{figure}

Object detection is one of the core tasks in computer vision. In recent years, a lot of  detectors based on ConvNets have been proposed to improve the accuracy and speed in object detection task \cite{Fast-RCNN,FasterRCNN,RFCN,SSD,YOLO2}. But scale variations in object detection is still a critical challenge for all of the detectors. As shown in Fig.~\ref{fig:fm}, there are some approaches which have been proposed to solve the multi-scale objects detection problem. Fig.~\ref{fig:fm} (a) applies a ConvNet to different scale images to generate different scale feature maps, which is a quite inefficient way. Fig.~\ref{fig:fm} (b) only selects one scale feature map but creates anchors with different scales to detect multi-scale objects. This method is adopted by Faster RCNN~\cite{FasterRCNN}, RFCN~\cite{RFCN} and so on. But the fixed receptive field size is a limitation to detect too large or too small objects. Top-Down structure like (c) in Fig.~\ref{fig:fm} is popular recently and has been proved working well in FPN~\cite{FPN}, DSSD~\cite{DSSD} and SharpMask~\cite{SharpMask}. But fusing features layer by layer is not efficient enough while there are many layers to be combined together.

The main trade-off of object detectors which are based on ConvNets is that the contradiction between object recognition and location. With deeper ConvNet, the feature maps can represent more semantic information with translation invariance, which is beneficial to object recognition but harmful to object location. To solve this problem, SSD adopts feature pyramid  to detect objects with different scales. For VGG16 \cite{VGG} as the backbone network, Conv4\_3 with feature stride 8 is used to detect small objects while Conv8\_2 with feature stride 64 to detect large objects. This strategy is rational because small objects will not lose too much location information in the shallow layers and large objects can also be  well located and recognized in the deeper layers. But the problem is that the feature of small objects generated by shallow layers lacks enough semantic information which will lead to the poor performance on small object detection. Besides, small objects also rely on the context information heavily \cite{small_RCNN}. The first row of Fig.~\ref{fig:det_results} shows us some missing detections for small objects by SSD.

In this paper, to tackle these problems mentioned above, we propose Feature Fusion SSD(FSSD) by adding a lightweight and efficient feature fusion module to the conventional SSD. We first define a framework of feature fusion module and abstract the factors which have critical effect on the feature fusion performance on object detection. According to the theoretic analysis and experiment results in Section~\ref{method} and Section~\ref{experiments}, the architecture of  feature fusion module is defined as follows: Features from different layers with different scales are projected and concatenated together followed by a Batch Normalization \cite{BN} layer to normalize the feature values. Then we append some down-sampling blocks to generate new feature pyramid, which are fed to multibox detectors to produce the final detection results. 

Using the proposed architecture, our FSSD improves a lot in performance at a slight expense of speed compared with conventional SSD. We evaluate the FSSD in VOC PASCAL~\cite{PASCAL_VOC} dataset and MSCOCO~\cite{COCO} dataset. The results indicate that our FSSD can improve the conventional SSD with a large margin especially for small objects without any bells and whistles. Besides, our FSSD also outperforms a lot of state-of-the-art object detectors based on VGGNet including ION \cite{inside-outside_2016} and Faster RCNN \cite{FasterRCNN}. Our feature fusion module can also  work better than FPN \cite{FPN} in the object detection task.
Our main contributions are summarized as follows:
\begin{itemize}
	\addtolength{\itemsep}{-0.1in}
	\item[(1)] We define the feature fusion framework and investigate the factors to confirm the structure of the feature fusion module.  
	\item[(2)] We introduce a novel and lightweight way of combining feature maps from different levels and generating feature pyramid to fully utilize the features.
	\item[(3)] With quantitative and qualitative experiments, we prove that our FSSD has a significant improvement over the conventional SSD with slight speed drop. FSSD can achieve state-of-the-art performance on both PASCAL VOC dataset and MS COCO dataset. 
\end{itemize}

\section{Related Work}
\paragraph {Object detector with deep ConvNet}
Benefited from the power of Deep ConvNes, object detector such as OverFeat \cite{overfeat} and R-CNN \cite{RCNN} have began to show the  dramatic improvements in accuracy. OverFeat applies a ConvNet as a feature extractor in the sliding window  on an image pyramid. R-CNN \cite{RCNN} uses the region proposals generated from selective search \cite{SS} or Edge boxes \cite{edgebox}  to generate the region-based feature from a pre-trained ConvNet and  SVMs are adopted to do classification.  SPPNet~\cite{SPPnet} adopts a spatial pyramid pooling layer which allows the classification module to reuse the ConvNet feature regardless of the input image resolutions. Fast R-CNN \cite{Fast-RCNN} introduces to train the ConvNet with both the classification and location regression loss end to end. Faster R-CNN \cite{FasterRCNN} suggests to replace selective search with a region proposal network (RPN). RPN is used to generate the candidate bounding boxes (anchor boxes) and filter out the background regions at the same time. Then another small network is used to do classification and bounding box location regression based on these proposals. R-FCN \cite{RFCN} replaces ROI pooling in the Faster RCNN with position sensitive ROI pooling (PSROI) to improve the detector's quality with both aspects of  accuracy and speed. Recently, Deformable Convolutional Network \cite{DCN} proposes deformable convolution and deformable PSROI to enhance the RFCN further with better accuracy.

Except the region based detectors, there are also some efficient one stage object detectors. YOLO (you only look once) \cite{YOLO} divides the input image into several grids and performs localization and classification on each part of image. Benefited from this method, YOLO can run object detection at a very high speed but the accuracy is not satisfactory enough. YOLOv2 \cite{YOLO2} is an enhanced version of YOLO and it improves the YOLO by removing the fully connected layers and adopts anchor boxes like the RPN.

\begin{figure*}[t]
	\centering
	\includegraphics[width=1\linewidth]{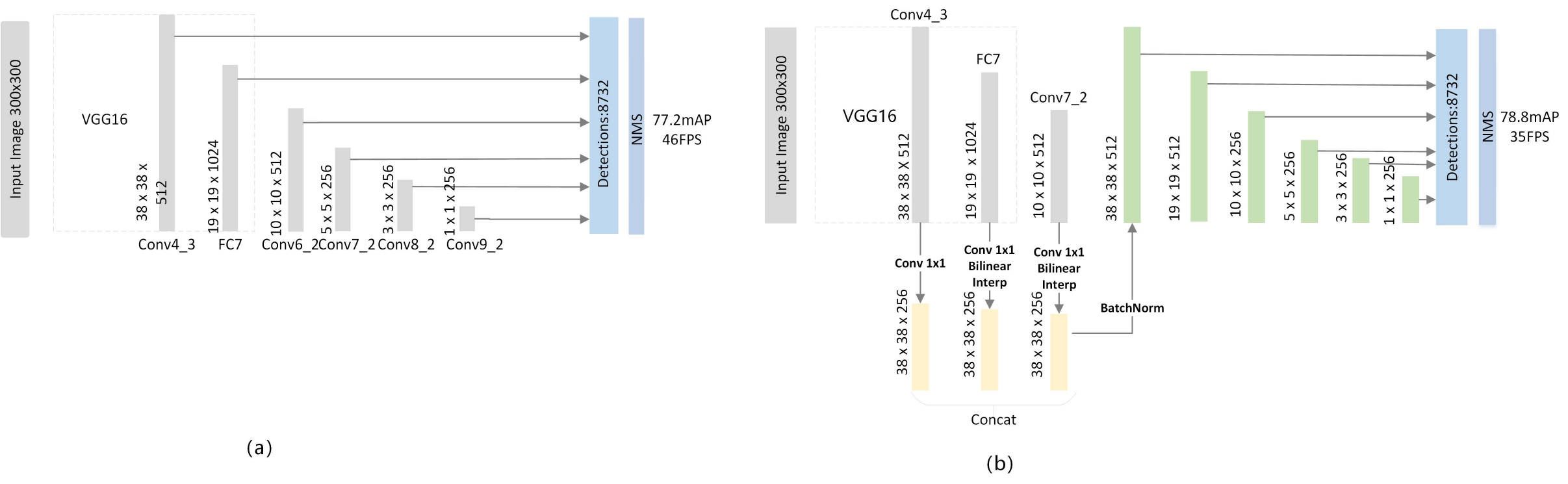}
	
	\caption{(a) is the SSD framework proposed in \cite{SSD}, (b) is our F-SSD framework.}
	\label{frame}
\end{figure*}

SSD \cite{SSD} is another efficient one stage object detector. As illustrated in Fig.~\ref{frame} (a), SSD predicts the class scores and location offsets for the default bounding boxes by two 3$\times$3 convolutional layers. In order to  detect objects with different scales, SSD adds a series of progressively smaller convolutional layers to generate pyramid feature maps and sets corresponding anchor size according to the receptive filed size of the layers. Then NMS (non-maximum suppression) is used to post-process  the final detection results. Because SSD detects objects directly from the plane ConvNet feature maps, it can achieve real-time object detection and process faster than most of other state-of-the-art object detectors. 

In order to improve the accuracy, DSSD \cite{DSSD} suggests to augment SSD+ResNet-101 with deconvolution layers to introduce additional larges-scale context. However, the speed is slow because of the model complexity. RSSD \cite{RSSD} uses rainbow concatenation through both pooling and concatenation to fully utilize the relationship between the layers in the feature pyramid to enhance the accuracy with a little speed lost. DSOD \cite{DSOD} investigates how to train a object detector from scratch and designs a DenseNet architecture to improve the parameter efficiency.
\paragraph { Algorithms using feature fusion in ConvNet}
There are a lot of approaches which attempt to use multiple layers' features to improve the performance in computer vision tasks. HyperNet \cite{hypernet:_2016}, Parsenet \cite{parsenet:_2015} and ION \cite{inside-outside_2016} concatenate features from multiple layers before predicting the result. FCN \cite{FCN}, U-Net \cite{unet}, Stacked Hourglass networks \cite{hourglass} also use skip connections to associate low-level and high-level feature maps to  fully utilize the  synthetic information. SharpeMask \cite{SharpMask} and FPN \cite{FPN} introduce top-down structure to combine the different level features together to enhance the performance. 
\label{feature_fusion}
\section{Method}
\label{method}
ConvNets have an excellent ability to extract pyramidal feature hierarchy, which has more semantic information  from low to high levels.  Conventional SSD regards these features in different levels as the same level and generates the object detection results directly from them. This strategy makes SSD lack the capability to capture both the local detailed features and global semantic features. However, the detector should merge the context information and their detailed features to confirm the small objects. So synthesizing the features with a slight structure is an important solution for a ConvNet object detector to improve the accuracy.  
\subsection{Feature Fusion Module}
As mentioned in Section \ref{feature_fusion}, there have been  a lot of algorithms trying to observe and fully utilize  the pyramidal features. The most common method is like Fig.~\ref{fig:fm} (c). This type of feature fusion is used in FPN \cite{FPN} and DSSD \cite{DSSD} and is verified to improve the conventional detector's performance a lot. But this design needs multiple feature merging processes. As shown in Fig.~\ref{fig:fm} (c), the new features on the right side can only fuse the features from the corresponding level features on the left and higher levels. Besides, latent feature and multiple features element-wise process also consume a lot of time. We propose a lightweight and efficient feature fusion module to handle this task. Our motivation is to fuse the different level features at once with a appropriate way and generate feature pyramid  from the fused features. There are several factors we should consider when design the feature fusion module. We will investigate them in the following part.
Assuming $X_i ,i \in \mathcal{C}$ are the source feature maps we want to fuse, the feature fusion module can be described as follows:

\begin{equation}
	X_f = \phi_{f}\{\mathcal{T}_i(X_i)\} \qquad i \in \mathcal{C}
\end{equation}
\begin{equation}
X^{'}_p = \phi_{p}(X_{f}) \qquad p \in \mathcal{P}
\end{equation}
\begin{equation}
loc,class = \phi_{c,l}(\cup\{{X^{'} _p\} )} \qquad p\in \mathcal{P}
\end{equation}
where  $\mathcal{T}_i$ means the transformation function of each source feature map before being concatenated together. $\phi_{f}$ is the feature fusion function. $\phi_{p}$ is the function to generate pyramid features. $\phi_{c,l}$ is the method to predict object detections from the provided feature maps.

We focus on the range of  layers which should be fused or not ($\mathcal{C}$), how to fuse the selected feature maps ($\mathcal{T}$ and $\phi_f$) and how to generate pyramid features ($\phi_{p}$). 

\paragraph{$\mathcal{C}$:}In the conventional SSD300 based on VGG16, the author chooses conv4\_3, fc\_7 of the VGG16 and new added layer conv6\_2, conv7\_2, conv8\_2, conv9\_2 to generate features to process object detections. The corresponding feature size is 38$\times$38, 19$\times$19, 10$\times$10, 5$\times$5, 3$\times$3 and 1$\times$1. We think that feature maps whose spatial size is smaller than 10$\times$10 have little information to merge, so we set the range of layers as conv3\_3, conv4\_3, fc\_7 and conv7\_2 ( we set the stride of conv6\_2 to 1 so the feature map size of conv7\_2 is 10$\times$10). According to the analysis in Section \ref{select_fea}, conv3\_3 brings no profit to the accuracy, so we do not fuse this layer. 

\paragraph{$\phi_{f}$:}There are mainly two ways to merge different feature maps together: concatenation and element-wise summation. Element-wise summation requires that feature maps should have the same size which means we have to convert the feature maps to the same channels. Because this requirement limits the flexibility of fusing feature maps, we prefer to use concatenation. Besides, according to the result in Section \ref{ff}, concatenation can get better result than element-wise summation. So we use concatenation to combine the features.

\paragraph{$\mathcal{T}$:}
In order to concatenate the features with different scales in a simple and efficient way, we adopt the following strategy. Conv 1$\times$1 is applied to each of the source layers to reduce the feature dimension firstly. Then we set the size of conv4\_3's feature map as the basic feature map size, which means our smallest feature step is 8. Feature maps generated from conv3\_3 are down-sampled to 38$\times$38 with 2$\times$2 max-pooling whose stride is 2. As for the feature maps whose size is smaller than 38$\times$38, we use bilinear interpolation to resize the feature maps to the same size with conv4\_3. In this way, all the features have the same size on spatial dimension. 
\paragraph{$\phi_{p}$:}
Adopted from the conventional SSD, we also use the pyramid feature map to generate object detection results. We test three different structures and compare the results to select the best one. According to the result in Section \ref{pf}, we select the structure which is composed of several simple blocks to extract the feature pyramid.
\begin{figure}[t]
	\centering
	\includegraphics[width=1\linewidth]{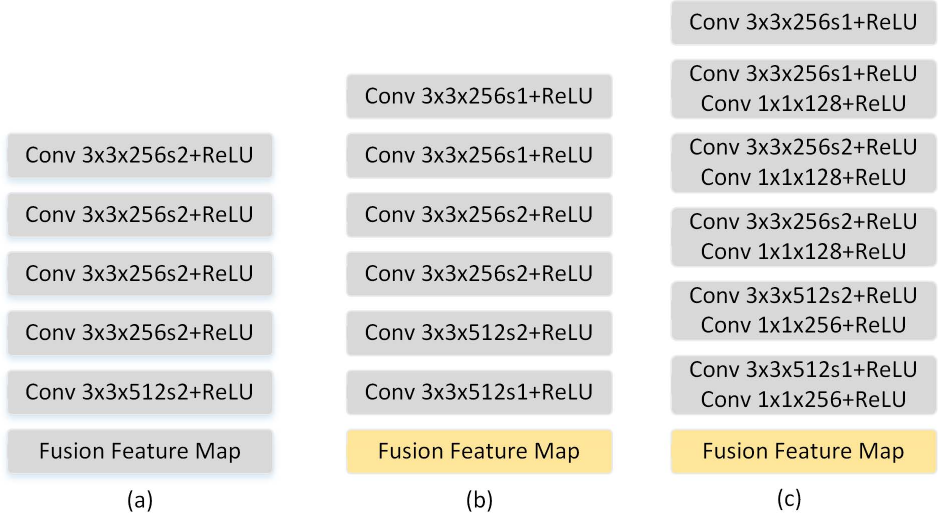}
	\caption{Pyramid feature generators for FSSD300. We use the feature maps from gray blobs to detect objects. In (a), the fusion feature map takes part in the object detection. In (b), we only detect objects on the feature maps after the fusion feature map. (c) We replace the simple group of Conv+ReLU with a bottleneck block which consists of  two Conv+ReLU layers. }
	\label{fig:pyramid}
\end{figure}
\setlength{\tabcolsep}{6pt}
\renewcommand{\arraystretch}{1.1}
\begin{table}[t]
	\begin{center}
		\footnotesize
		\begin{tabular}{c|c}
			& VOC2007 \texttt{test}\\\hline
			structure & mAP \\
			\hline
			simple block including fusion feature 	& 78.2 			\\
			simple block after fusion feature	& \textbf{78.8}					\\
		    bottleneck block after fusion feature	& 78.4					\\\hline
		\end{tabular}
	\end{center}
	\caption{mAP on VOC2007 \texttt{test} while using different pyramid feature generation structure.
	}\vspace{-3mm}
	\label{tab:prymid}
\end{table}


\subsection{Training}
\begin{figure}[t]
	\centering
	\includegraphics[width=1\linewidth]{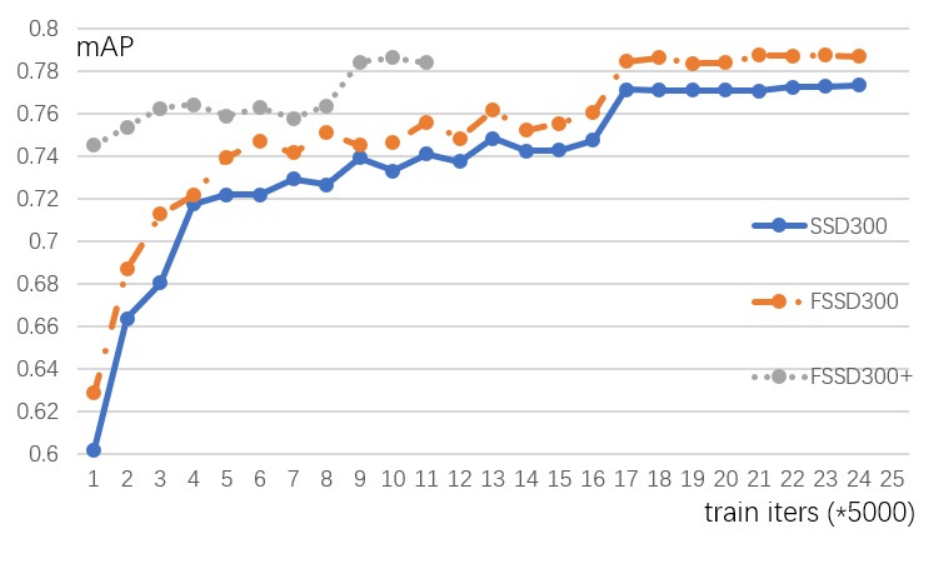}
	\caption{Training process comparison. The vertical axis denotes the mAP calculated on VOC2007 \texttt{test} set and the horizontal axis represents the training iterations. SSD means that training the conventional SSD model with the default settings from a pre-trained VGG16 model. FSSD means that we train the FSSD model with a pre-trained VGG model. FSSD's training parameters are the same with SSD. FSSD+ means that we train the FSSD from a pre-trained SSD model. The FSSD+ is only optimized for 60k iterations. All of the models are trained on VOC07+12 dataset.}
	\label{fig:train}
\end{figure}

There are two main training methods we can choose. Firstly, because our FSSD is based on the SSD model, we can adopt the well trained SSD model as our pre-trained model.  As for the learning rate, we set the new feature fusion module's learning rate two times larger than other parameters. Another way to train the FSSD is the same as the conventional SSD algorithm. According to the experiments in Table \ref{ablation}(rows 2 and 5), these two ways have little difference on the final result but training FSSD from VGG16 has a slightly better results than the one which is trained from SSD model. But as illustrated in Fig.~\ref{fig:train}, training FSSD from SSD models can converge faster than from the pre-trained VGGNet model. Besides, our FSSD also  coverages faster than the conventional SSD while having the same training hyper-parameters . In order to test our algorithms in the  limited time, we use the first way to train the FSSD by default.

The training objective is the same as SSD. We use the center code type to encode the bounding boxes and have the same matching strategy, hard negative mining strategy and data augmentation with SSD. More details can be seen in \cite{SSD}.

\section{Experiments}
\label{experiments}
In order to compare our FSSD with the conventional SSD fairly, our experiments are all based on VGG16\cite{VGG} which is preprocessed like SSD~\cite{SSD}. We conduct experiments on PASCAL VOC 2007, 2012 \cite{PASCAL_VOC} and MS COCO dataset \cite{COCO}. In VOC 2007 and VOC 2012, a predicted bounding box is correct if its intersection over union (IOU) with the ground truth is higher than 0.5. We adopt the mean average precision (mAP) as the metric for evaluating detection performance. As for MSCOCO, we upload the result to the evaluation server to get the performance analysis. All of our experiments are based on the Caffe~\cite{caffe} version of  SSD implementation.
\subsection{Ablation Study on PASCAL VOC2007}
\label{ablation_section}
\begin{table*}\small
	\begin{center}
		\resizebox{1\textwidth}{!}{%
		\begin{tabular}{c|c|c|c|c|c|c|c}
			\hline
			Method     & data  & BN  & pre-trained VGG &  pre-trained SSD & feature fusion & fusion layers& mAP  \\ \hline
			FSSD300   & 07+12 &\Checkmark&\ding{55}& \ding{55} & Concat&conv3-conv7& 72.7 \\
			FSSD300   & 07+12 &\Checkmark&\ding{55}& \Checkmark & Concat&conv3-conv7& 78.6 \\
			FSSD300   & 07+12 &\Checkmark&\ding{55}& \Checkmark & Concat&conv4-conv7& \textbf{78.8} \\
			FSSD300   & 07+12 &\Checkmark&\ding{55}& \Checkmark & Concat&conv4-fc7& 78.6 \\
			FSSD300   & 07+12 &\Checkmark&\Checkmark& \ding{55} & Concat&conv3-conv7& \textbf{78.8} \\
			
			FSSD300   & 07+12 &\ding{55}&\ding{55}& \Checkmark & Concat&conv3-conv7& 78.1 \\
			FSSD300   & 07+12 &\Checkmark&\ding{55}& \Checkmark & ele-sum&conv3-conv7& 76.3 \\
			
			\hline  
		\end{tabular}
	}
	\end{center}
	\caption{Results of the ablation study on PASCAL VOC2007. \textbf{BN} means that a Batch Normalization layer is added after the feature concatenation. \textbf{pre-trained VGG} means that a pre-trained VGG16 is adopted to initialize the model. \textbf{pre-trained SSD} means that the FSSD is optimized from a well-trained SSD model. The options of layers we can fuse include \textit{conv3\_3, conv4\_3 ,fc\_7, conv7\_2}. The \textbf{fusion layers} represents the layers we choose to merge. The mAP is measured on VOC2007 \texttt{test} set. }
	\label{ablation}
\end{table*}

In this Section, we investigate some important design factors about the feature fusion module. We compare the results on PASCAL VOC 2007 with input size 300$\times$300. In these experiments, the models are trained with the combined dataset from 2007 \texttt{trainval} and 2012 \texttt{trainval} (VOC07+12) and tested on VOC 2007 \texttt{test} set. The results are summarized in Table \ref{ablation} mostly.
\subsubsection{The range of layers to be fused}
\label{select_fea}
In Table~\ref{ablation}, we compare the FSSD with different fusion layers. While we fuse all of the feature maps (conv3\_3, conv4\_3, fc\_7 and conv7\_2), the mAP on VOC2007 \texttt{test} set(row 2) is 78.6\%. It is interesting that if we remove the conv3\_3, the mAP is increased to 78.8\%(row 5), which means that the down-sampled feature map from conv3\_3 whose feature stride is 4 has no benefit to the final performance. But according to the result in Table~\ref{ablation} 4th row, it is better to keep the conv7\_2.
\subsubsection{Feature fusion: concatenation or element-wise summation}
\label{ff}
 In Table \ref{ablation}, using concatenation to fuse feature can get 78.6\% mAP(row 2) while element-wise summation can only achieve 76.3\%(row 7). The results show that concatenation is better than element-wise summation with a large margin of 2.3 points.
\subsubsection{Normalize feature value or not}
Feature maps from different layers always have different ranges of value. In conventional SSD, the author used L2Normalization to scale the feature map from conv4\_3. As we want to use a simple and efficient way to scale the feature maps, we add a Batch Normalization layer after the concatenation process. Results in Table \ref{ablation}(rows 2 and 6) show that using the Batch Normalization layer to re-scale the feature maps can bring us about 0.7\% mAP improvement.
\subsubsection{Pyramid feature extractor design}
\label{pf}
We compare three different structures of pyramid feature extractor as shown in Fig.~\ref{fig:pyramid}. There are two kinds of blocks to generate lower resolution feature maps, simple block (one Conv$3\times3$ followed by a ReLU) and bottleneck block, which uses Conv $1\times1$ to reduce the feature dimension first and is adopted by the conventional SSD. Structure in  Fig.~\ref{fig:pyramid} (a) is composed of several simple blocks and uses the fused feature to predict objects. Fig.~\ref{fig:pyramid} (b) is the same with (a) except not using fused feature for prediction directly. Fig.~\ref{fig:pyramid} (c) replaces the simple block with bottleneck block in (b). The results which are shown in Table~\ref{tab:prymid} indicate that structure (b) is better than other two designs with a little improvement. 
\subsection{Results on PASCAL VOC}
\begin{table*}[]
	\centering
	\resizebox{1\textwidth}{!}{%
		\begin{tabular}{c|c|c|c|c|c|c|c|c}
			\hline
			Method     & data  & pre-train &    backbone network  & speed (\textit{fps}) & GPU &\#proposals&input size & mAP  \\ \hline
			Faster RCNN~\cite{FasterRCNN}  & 07+12 &\Checkmark&     VGGNet    &     7    & Titan X &6000& $\sim600\times1000$&73.2 \\
			Faster RCNN~\cite{FasterRCNN}  & 07+12 &\Checkmark&   ResNet-101    &     2.4      & K40&300 &$\sim600\times1000$& 76.4 \\
			R-FCN~\cite{RFCN}     & 07+12 &\Checkmark&   ResNet-50    &     -      & -& 300&$\sim600\times1000$& 77.0 \\
			R-FCN~\cite{RFCN}     & 07+12 &\Checkmark&   ResNet-101    &     5.8      & K40& 300&$\sim600\times1000$& 79.5 \\  \hline
			YOLOv2~\cite{YOLO2}   & 07+12 &\Checkmark&  Darknet-19       &       81  &Titan X&-&$352\times352$&  73.7 \\ \hline
			SSD300S$^\dagger$~\cite{DSOD}&07+12&\ding{55}& VGGNet&46&Titan X &8732&$300\times300$&69.6\\
			SSD300*~\cite{SSD}   & 07+12+COCO &\Checkmark&  VGGNet       &      46  &Titan X&8732&$300\times300$&  81.2 \\
			SSD300~\cite{SSD}   & 07+12 &\Checkmark&   VGGNet       &    46   &Titan X&8732& $300\times300$& 77.2 \\
			SSD300~\cite{SSD}   & 07+12 &\Checkmark&   VGGNet       &    85   &1080Ti&8732& $300\times300$& 77.2 \\
			SSD512~\cite{SSD}    & 07+12 &\Checkmark&    VGGNet       &      19   &Titan X&24564&$512\times512$&  78.5 \\
			SSD512*~\cite{SSD}   & 07+12+COCO &\Checkmark&    VGGNet       &     19   &Titan X&24564&$512\times512$&  83.2 \\
			\hline
			DSOD300~\cite{DSOD}    & 07+12 &\ding{55}&  DS/64-192-48-1  &     17.4    &  Titan X& -&$300\times300$&  77.7 \\
			DSOD300*~\cite{DSOD}    & 07+12+COCO &\ding{55}&  DS/64-192-48-1  &   17.4  &  Titan X& -&$300\times300$& 81.7 \\ \hline
			DSSD321~\cite{DSSD}    & 07+12&\Checkmark & ResNet-101 &9.5&Titan X&17080&$321\times321$&78.6\\
			DSSD513~\cite{DSSD}    & 07+12&\Checkmark & ResNet-101 &5.5&Titan X&43688&$321\times321$&81.5\\ \hline
			RSSD300~\cite{RSSD}    & 07+12&\Checkmark &  VGGNet &35&Titan X&8732&$300\times300$&78.5\\
			RSSD512~\cite{RSSD}    & 07+12&\Checkmark &  VGGNet &16.6&Titan X&24564&$512\times512$&80.8\\ \hline
			FSSD300S$^\dagger$               &07+12 &\ding{55}&        VGGNet    & 65.8  &1080Ti&8732&$300\times300$&72.7\\
			FSSD300                &07+12 &\Checkmark&        VGGNet    & 65.8  &1080Ti&8732&$300\times300$&78.8\\
			FSSD300*               &07+12+COCO &\Checkmark&        VGGNet    & 65.8  &1080Ti&8732&$300\times300$&\textbf{82.7}\\
			FSSD512                &07+12 &\Checkmark &       VGGNet    & 35.7  &1080Ti&24564&$512\times512$&80.9\\   
			FSSD512*                &07+12+COCO &\Checkmark &       VGGNet    & 35.7  &1080Ti&24564&$512\times512$&\textbf{84.5}\\      
			\hline
		\end{tabular}
	}
	\caption{\textbf{PASCAL VOC 2007 \texttt{test} detection results.} SSDs' results we cited here are the updated version by the author after the paper publication with more data argumentation. SSD300S$^\dagger$ indicates training SSD300 from scratch VGGNet, which is tested in DSOD\cite{DSOD}. FSSD300S$^\dagger$ is also trained from scratch VGGNet. The speed of FSSDs is tested on a single Nvidia 1080Ti GPU. For fair comparison, we also test the SSD's speed on the same single Nvidia 1080Ti GPU. } 
	\label{VOC2007}
	\end{table*}
\begin{table*}[]\small
	\centering
	\setlength{\tabcolsep}{1.82pt}
	\resizebox{1\textwidth}{!}{%
		\begin{tabular}{l|c|c|c|c|cccccccccccccccccccc}
			\hline
			Method &  data  &  backbone network &  pre-train &  mAP &  aero &  bike &  bird &  boat &  bottle &  bus &  car &  cat &  chair &  cow &  table &  dog &  horse &  mbike &  person &  plant &  sheep &  sofa &  train &  tv \\
			\hline
			Faster RCNN~\cite{FasterRCNN} &  07++12+COCO &  ResNet-101 & \Checkmark& 83.8&	92.1&	88.4&	84.8&	\textbf{75.9}&	\textbf{71.4}&	86.3&	87.8&	94.2&	66.8&	89.4&	69.2&	\textbf{93.9}&	91.9&	90.9&	89.6&	\textbf{67.9}&	88.2&	76.8&	90.3&	80.0 \\
		    Faster RCNN~\cite{FasterRCNN} &  07++12+COCO& VGGNet & \Checkmark&75.9&	87.4&	83.6&	76.8&	62.9&	59.6&	81.9&	82.0&	91.3&	54.9&	82.6&	59.0&	89.0&	85.5&	84.7&	84.1&	52.2&	78.9&	65.5&	85.4&	70.2 \\ \hline
			YOLOv2~\cite{YOLO2} &  07++12+COCO &  Darknet-19 & \Checkmark& 81.5&	90.0&	88.6&	82.2&	71.7&	65.5&	85.5&	84.2&	92.9&	67.2&	87.6&	70.0&	91.2&	90.5&	90.0&	88.6&	62.5&	83.8&	70.7&	88.8&	79.4	 \\
			BlitzNet512~\cite{blitznet}&07++12+S+COCO&ResNet-50& \Checkmark&83.8&	\textbf{93.1}&	89.4&	84.7&	75.5&	65.0&	86.6&	87.4&	94.5&	\textbf{69.9}&	88.8&	\textbf{71.7}&	92.5&	91.6&	91.1&	88.9&	61.2&	\textbf{90.4}&	\textbf{79.2}&	\textbf{91.8}&	\textbf{83.0}	\\
			SSD300~\cite{SSD} &  07++12+COCO&
			VGGNet & \Checkmark&79.3&	91.0&	86.0&	78.1&	65.0&	55.4&	84.9&	84.0&	93.4&	62.1&	83.6&	67.3&	91.3&	88.9&	88.6&	85.6	&54.7&	83.8&	77.3&	88.3&	76.5\\	
			SSD512~\cite{SSD}&07++12+COCO&
			VGGNet& \Checkmark&82.2	&91.4&	88.6&	82.6&	71.4&	63.1&	87.4&	88.1&	93.9&	66.9&	86.6&	66.3&	92.0&	91.7&	90.8&	88.5&	60.9&	87.0&	75.4&	90.2&	80.4\\	
			DSOD300~\cite{DSOD} &  07++12+COCO&
			DS/64-192-48-1 & \ding{55} & 79.3 &  90.5 & 87.4 & 77.5 & 67.4 & 57.7 & 84.7 & 83.6 & 92.6 & 64.8 & 81.3 & 66.4 & 90.1 & 87.8 & 88.1 & 87.3 & 57.9 & 80.3 & 75.6 & 88.1 & 76.7 \\
			\hline
				FSSD300 &  07++12+COCO&
		VGGNet & \Checkmark &\textbf{82.0}&92.2&	89.2&	81.8&	72.3&	59.7&	87.4&	84.4&	93.5&	66.8&	87.7&	70.4&	92.1&	90.9&	89.6&	87.7&	56.9&	86.8&	79.0	&90.7&	81.3	 \\
				FSSD512 &  07++12+COCO&
		VGGNet & \Checkmark & \textbf{84.2}&	92.8&	\textbf{90.0}&	\textbf{86.2}&	\textbf{75.9}&	67.7&	\textbf{88.9}&	\textbf{89.0}&	\textbf{95.0}&	68.8&	\textbf{90.9}&	68.7&	92.8&	\textbf{92.1}&	\textbf{91.4}&	\textbf{90.2}&	63.1&	90.1&	76.9&	91.5&	82.7\\\hline
	
		\end{tabular}
	}

	\caption{\textbf{PASCAL VOC 2012 \texttt{test} detection results.}
		\textbf{07++12+COCO}: 07 \texttt{trainval} + 07 \texttt{test} + 12 \texttt{trainval} + MSCOCO. \textbf{07++12+S+COCO}: 07++12 plus segmentation labels and MSCOCO. Result links are FSSD300 (07++12+COCO) : \url{http://host.robots.ox.ac.uk:8080/anonymous/YMA3TZ.html}; FSSD512 (07++12+COCO): \url{http://host.robots.ox.ac.uk:8080/anonymous/LQXCQK.html}.}
\label{VOC2012}
\end{table*}
\paragraph{Experimental setup}
According to the ablation study in Section \ref{ablation_section}, the architecture of our FSSD is defined as follows: 
For FSSD with $300\times300$ input(FSSD300), we adopt VGG16 as the backbone network. All of the raw features are converted to 256 channels with a $1\times1$ convolutional layer. Feature maps from fc\_7 and conv7\_2 are interpolated to $38\times38$. Then the transformed feature maps are concatenated together followed by a Batch Normalization layer to normalize the feature values. Then several down-sampling blocks(including one $3\times3$ convolutional layer with stride 2 and one ReLU layer ) are appended one by one to generate the pyramid features.
\paragraph{Results on PASCAL VOC2007}
We use VOC 2007 \texttt{trainval} and VOC2012 \texttt{trainval} to train FSSD following SSD~\cite{SSD}.  We train the FSSD300 on two Nvidia 1080Ti GPUs with batch size 32 for 120k iterations. The initial learning rate is set to 0.001 and then divided by 10 at step 80k, 100k and 120k. Following the training strategy in SSD~\cite{SSD}, the weight decay is set to 0.0005. We adopt a SGD with momentum 0.9 to optimize the FSSD which is initialized by a well pre-trained VGG16 on ImageNet. In order to use COCO models as the pre-trained model, we first train the COCO model  with 80 classes which will be described in details in Section \ref{COCO_result} . Then a subset of the 80 classes model which is corresponding to the classes in PASCAL VOC 20 classes are extracted from the COCO model as the pre-trained VOC model.
Our results on VOC2007 \texttt{test} set are shown in Table \ref{VOC2007}. Our FSSD300 can achieve 78.8\%  mAP, which improves 1.1 points compared with the conventional SSD300. In addition, our result of FSSD300  is also higher than DSSD321, even though DSSD321 uses ResNet-101 as the backbone network, which has better performance compared with VGG16.  With COCO as the training data, FSSD300's performance can be further increased to 82.7\%, which exceeds DSOD by 1\%  and SSD300 by 1.5\%. Our FSSD512 also improves the SSD512 from 79.8\% to 80.9\%, which is also a little higher than RSSD512. DSSD512 is better than our FSSD512 with the aspect of accuracy but we think the ResNet-101 backbone plays a critical role in this progress. However, our FSSD512 is much faster than DSSD512. 

It is interesting that DSOD~\cite{DSOD} researches the object detections task with a particular way: training an object detector from scratch. The conventional SSD with VGG16 can only achieve 69.6\% mAP without the pre-trained VGGNet model. We also investigate whether our FSSD can improve the conventional SSD. Result in Table \ref{VOC2007} (row 19) shows us that our FSSD trained from scratch also improves the performance with a large margin 3.1 points. Even though this result is not as good as the DSOD, it should be noted that our FSSD still takes VGGNet as the base network while DSOD pays more attention on the design of the backbone network for better performance and DSOD is trained for more iterations (about 4 times than SSD).
\paragraph{Results on PASCAL VOC2012}
We use VOC 2012 \texttt{trainval}, VOC 2007 \texttt{trainval\_test} and MSCOCO to train our FSSD and test it on VOC2012 \texttt{test} set. Training hyper-parameters are the same with the experiments on VOC2007 except the training dataset. Table \ref{VOC2012} summarizes some state-of-the-art detectors from the PASCAL VOC2012 leaderboard. FSSD300 with COCO can achieve 82.0\% mAP, which is higher than the conventional SSD(79.3\%) by 2.7 points. In addition, our FSSD512 can achieve 84.2\% mAP, which exceeds the conventional SSD(82.2\%) by 2 points. The FSSD512 takes the first place in VOC2012 leaderboard among all of the one-stage object detectors  as of the time of submission.
\subsection{Results on MS COCO}
\label{COCO_result}
\begin{table*}[]
	\centering
	\resizebox{1\textwidth}{!}{%
		\begin{tabular}{l|c|c|c|ccc|ccc|ccc|ccc}
			\hline
			\multirow{2}{*}{Method}          & \multirow{2}{*}{data} & \multirow{2}{*}{network} & \multirow{2}{*}{pre-train} & \multicolumn{3}{c|}{Avg. Precision, IoU:} & \multicolumn{3}{c|}{Avg. Precision, Area:} & \multicolumn{3}{c|}{Avg. Recall, \#Dets:} & \multicolumn{3}{c}{Avg. Recall, Area:} \\
			&                       &              &            & 0.5:0.95        & 0.5        & 0.75       & S            & M            & L            & 1            & 10           & 100         & S           & M           & L           \\ \hline
			Faster RCNN~\cite{FasterRCNN} & trainval              & VGGNet          &    \Checkmark     & 21.9            & 42.7       & -          & -            & -            & -            & -            & -            & -           & -           & -           & -           \\
			ION~\cite{inside-outside_2016}  & train                 & VGGNet    &    \Checkmark   & 23.6            & 43.2       & 23.6       & 6.4          & 24.1         & 38.3         & 23.2         & 32.7         & 33.5        & 10.1        & 37.7        & 53.6        \\
			R-FCN~\cite{RFCN}        & trainval              & ResNet-101 & \Checkmark & 29.2            & 51.5       & -          & 10.3         & 32.4         & 43.3         & -            & -            & -           & -           & -           & -           \\
			R-FCN{\footnotesize{multi-sc}}~\cite{RFCN}        & trainval              & ResNet-101 & \Checkmark & 29.9            & 51.9       & -          & 10.8         & 32.8         & 45.0         & -            & -            & -           & -           & -           & -           \\ \hline
			YOLOv2~\cite{YOLO2}        & trainval35k           & Darknet-19   & \Checkmark      & 21.6            & 44.0       & 19.2       & 5.0          & 22.4         & 35.5         & 20.7         & 31.6         & 33.3        & 9.8        & 36.5        & 54.4        \\ 		
			SSD300*~\cite{SSD}        & trainval35k           & VGGNet   & \Checkmark      & 25.1            & 43.1       & 25.8       & 6.6          & 25.9         & 41.4         & 23.7         & 35.1         & 37.2        & 11.2        & 40.4        & 58.4        \\
			SSD512*~\cite{SSD}        & trainval35k           & VGGNet   & \Checkmark      & 28.8            & 48.5   &30.3   & 10.9      & 31.8         & 43.5        & 26.1               & 39.5       & 42.0       &16.5       & 46.6       & 60.8        \\
			DSOD300~\cite{DSOD}                           & trainval         & DS/64-192-48-1 & \ding{55} & 29.3            & 47.3       & 30.6       & 9.4          &   31.5         & 47.0         & 27.3         & 40.7         & 43.0        & 16.7        & 47.1        & 65.0        \\
			
			DSSD321~\cite{DSSD}                           & trainval35k         & ResNet-101 & \Checkmark & 28.0            & 46.1       & 29.2       & 7.4          & 28.1         & 47.6         & 25.5         & 37.1         & 39.4        & 12.7        & 42.0        & 62.6     \\  
            DSSD513~\cite{DSSD} & trainval35k         & ResNet-101 & \Checkmark& \textbf{33.2}           &\textbf{53.3}      & \textbf{35.2}     & 13.0         & \textbf{35.4}        & \textbf{51.1}         & \textbf{28.9}         &\textbf{43.5}         & \textbf{46.2}        & 21.8        & 49.1        & \textbf{66.4}       \\\hline
			FSSD300                           & trainval35k         & VGGNet &\Checkmark & 27.1           & 47.7       & 27.8       & 8.7          & 29.2         & 42.2         & 24.6         & 37.4         & 40.0        & 15.9        & 44.2        & 58.6        \\ 
			FSSD512                           & trainval35k         & VGGNet & \Checkmark & 31.8           & 52.8      &33.5       & \textbf{14.2}          & 35.1         & 45.0         & 27.6     &    42.4         & 45.0        & \textbf{22.3}       &\textbf{49.9}        & 62.0        \\ \hline
		\end{tabular}
	}
	\caption{\textbf{MS COCO \texttt{test-dev 2015} detection results.}}
	\label{COCO}
\end{table*}
MSCOCO dataset has 80 object categories. We use the COCO Challenge 2017 data split to prepare our dataset. The train set involves 115K images which is comparable with the original trainval35k. We test the FSSDs on the 20k test-dev set. For training FSSD300, the learning rate is set to 0.001 for the first 280k iterations, then divided by 10 at the step 360k and 400k. But if we train the FSSD300 from a well trained SSD model, it will only need 120k totally to make the FSSD300 converge well. For training FSSD512, the learning rate is set to 0.001 for the first 280k iterations and then divided by 10 at the step 320k and 360k. 

The COCO test results are shown in Table \ref{COCO}. FSSD 300 achieves 27.1\% on the  {\texttt {test-dev}} set, which is higher than the SSD300*(25.1\%) with a large margin. Even though our FSSD does not perform as well as DSOD and DSSD, it should be noted that our base model is VGG16 and FSSD has the best accuracy compared with other algorithms such as  Faster RCNN and ION in Table \ref{COCO}(rows 1 and 2)  which is also based on VGGNet. Besides, FSSD512(31.8\%) outperforms conventional SSD(28.8\%) by 3 points. Even though our FSSD512 is slightly lower than DSSD513, it should be noted that FSSD's mAP on small objects is still higher than DSSD513, which proves that our feature fusion module is more powerful than DSSD's FPN module on small objects' detection.

\subsection{Speed }
\begin{figure}[t]
 
	\centering
	\includegraphics[width=1\linewidth]{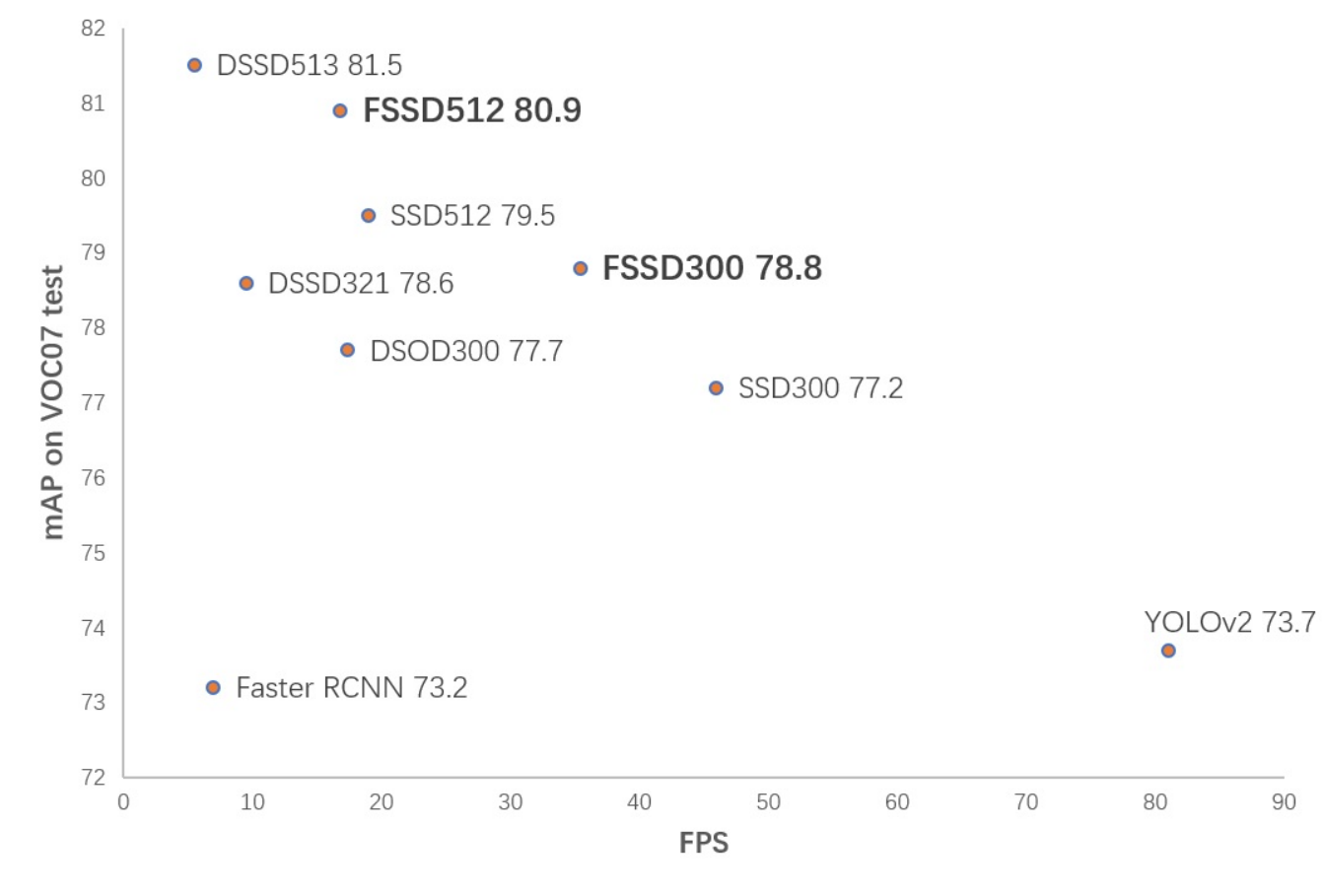}

	\caption{Speed and  accuracy distribution with different object detection algorithms. The speeds are all measured on a single Titan X GPU. As we do not have a Titan X GPU, the speed of our FSSDs is calculated by comparing with SSD's speed which we have tested on our own Nvidia 1080Ti.}
	  \label{fig:speed}
\end{figure}
The inference speed is shown in the 5th column of Table \ref{VOC2007}. Our FSSD can run at 65.8 FPS with a $300\times300$ input image on a single 1080Ti GPU. For fair comparisons, we also test the speed of SSD on the 1080Ti GPU. Because our FSSD adds additional layers on the SSD model, our FSSD consumes about 25\% extra time. But compared with DSSD and DSOD, our methods  are still much faster than them while the improvement from the SSD is about the same level. In Fig.~\ref{fig:speed}, it is clear that our FSSD is faster than most of the object detection algorithms while having the competitive accuracy.
\subsection{Performance improvement from SSD}
\begin{figure*}[t]
	\centering
	\includegraphics[width=1\linewidth]{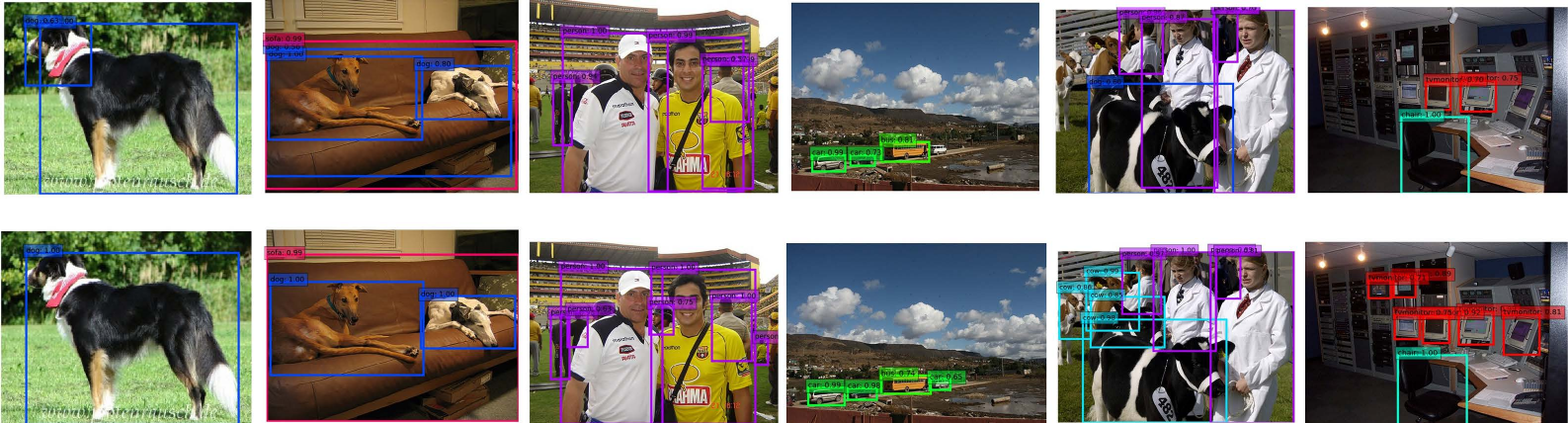}
	\caption{SSD300 vs FSSD300. Both models are trained with VOC07+12 dataset. The top row contains the results from the conventional SSD300 and the bottom row is from our FSSD300 model. Bounding boxes with score of 0.5 or higher is drawn. Better viewed on screen. }	
	\label{fig:det_results}

\end{figure*}
Our FSSD performs better than conventional SSD mainly in two aspects. Firstly, FSSD reduces the probability of detecting multi-part of one object repeatedly or merging multi-objects to one single object . For example, as illustrated in Fig.~\ref{fig:det_results} column 1, The SSD model regards the head of the dog as a single dog as well, which is clearly wrong while the whole dog is in this image. However, FSSD can detect the whole dog at once. In addition, the column 2 in Fig.~\ref{fig:det_results} shows us that the SSD detects one big dog which includes two dogs in fact. But FSSD does not make the mistake. Secondly, FSSD performs better on small objects. On the one hand, small objects can only activate smaller regions in the network comparing with large objects and the location information is easy to be lost in the detection process. On the other hand, small object's recognition relies more on the context around it. Because SSD only detects small objects from the shallow layers such as conv4\_3, whose receptive field is too small to observe the lager context information, which leads to the SSD's bad performance on small objects. FSSD can observe all of the objects synthetically benefited from the feature fusion module. As shown in Fig.~\ref{fig:det_results} column 3 to column 6. FSSD detects more small objects than SSD successfully.

\section{Conclusion and future work}
In this paper, We proposed FSSD, an enhanced SSD by applying a lightweight and efficient feature fusion module on it. First, we investigate the framework to fuse different features together and generate pyramid feature maps. The experiments' results show that feature maps from different layers can be well fully utilized by being concatenated together. Then several convolutional layers with stride 2 are applied to fused feature map to generate pyramid features. Experiments on VOC PASCAL and MSCOCO prove that our FSSD improves the conventional SSD a lot and out-performs several other state-of-the-art object detectors both in accuracy and efficiency without any bells and whistles.
 
In the future, it is worth enhancing our FSSD with much stronger backbone networks such as ResNet~\cite{resnet} and DenseNet~\cite{Densenet} to get better performance on the MSCOCO dataset and replacing the FPN in Mask RCNN \cite{MaskRCNN} with our feature fusion module is also an interesting research field.
{\small
\bibliographystyle{ieee}
\bibliography{FSSD}
}

\end{document}